\documentclass[runningheads]{llncs}
\usepackage[T1]{fontenc}
\usepackage{graphicx,verbatim}
\usepackage{graphicx}%
\usepackage{multirow}%
\usepackage{amsmath,amssymb,amsfonts}%
\usepackage{mathrsfs}%
\usepackage[title]{appendix}%
\usepackage{xcolor}%
\usepackage{textcomp}%
\usepackage{manyfoot}%
\usepackage{booktabs}%
\usepackage{algorithm}%
\usepackage{algorithmicx}%
\usepackage{algpseudocode}%
\usepackage{listings}%
\usepackage{adjustbox}
\usepackage{booktabs}
\usepackage{pifont} %https://satztexnik.com/tex-archive/macros/latex/required/psnfss/psnfss2e.pdf
\usepackage{siunitx}
\usepackage{placeins}
\usepackage{hyperref}
\usepackage{amsmath}
\usepackage{mathrsfs}
\usepackage{upgreek}
\usepackage{nicematrix}
\usepackage{siunitx}

\usepackage{esvect}

\newcommand{\task}{\mathcal{T}}

\usepackage{colortbl}
\definecolor{tabbestcolor}{rgb}{0.004, 0.141, 0.337}
\definecolor{hotcolor}{rgb}{1.0, 0.666, 0.666}

\def \best {\cellcolor{tabbestcolor!30}}
\def \sbest {\cellcolor{tabbestcolor!15}}
\def \hot {\cellcolor{hotcolor!30}}

\definecolor{red_bright}{HTML}{faf0f0}
\definecolor{red_dark}{HTML}{e5bab9}
\definecolor{green_bright}{HTML}{e2feee}
\definecolor{green_dark}{HTML}{a6dfbf}
\definecolor{blue_bright}{HTML}{deeefd}
\definecolor{blue_dark}{HTML}{afc4db}

\algdef{SE}[SUBALG]{Indent}{EndIndent}{}{\algorithmicend\ }%
\algtext*{Indent}
\algtext*{EndIndent}

\begin{document}
\title{Sterilizable Scene Graph Generation\\ for Operating Rooms}

\author{Nick Lemke\inst{1,2} \and Ssharvien Kumar Sivakumar\inst{1,3} \and Antoine P. Sanner\inst{4} \and John Kalkhof\inst{5} \and Henry John Krumb\inst{1} \and Ghazal Ghazaei\inst{3} \and Anirban Mukhopadhyay\inst{1}}
\authorrunning{N. Lemke et al.}
\institute{Technical University of Darmstadt, Darmstadt, Germany \\
\email{nick.lemke@gris.informatik.tu-darmstadt.de}
\and ImFusion GmbH, Munich, Germany
\and Carl Zeiss AG, Munich, Germany 
\and University Medical Center Mainz, Mainz, Germany
\and Inria Center at University Côte d'Azur, Sophia Antipolis, France
}

\maketitle              % typeset the header of the contribution
\begin{abstract}
Scene graph generation from surgical video enables a holistic and structured understanding of surgical scenes by modeling objects and their semantic relationships.
Despite recent advances, state-of-the-art approaches rely on large, parameter-heavy deep learning models that are impractical for deployment in the operating room (OR) due to hardware footprint, hygiene constraints, latency, and data privacy concerns.
To the best of our knowledge, this is the first scene graph generation method built on NCAs and the first NCA framework capable of learning structured representations.
We introduce SG-NCA, a lightweight scene graph generation framework based on Neural Cellular Automata (NCA), designed for inference in fanless devices critical for OR hygiene protocols.
SG-NCA is the first scene graph generation combining NCA-based multi-class segmentation for efficient object detection and feature extraction with a lightweight relation predictor.
We evaluate SG-NCA on videos of cataract surgery and cholecystectomy, demonstrating performance comparable to established baselines while requiring 55× fewer parameters. 
We showcase deployment on fanless edge devices better suited for the OR and demonstrate downstream applications such as surgical video captioning, highlighting SG-NCA's potential for affordable, privacy-preserving, and OR-ready intraoperative scene understanding.
Our code is publicly available at: \url{https://github.com/MECLabTUDA/SG-NCA}
% Authors must provide keywords and are not allowed to remove this Keyword section.

\end{abstract}

\section{Introduction}
\begin{figure}[t]
    \centering
    \includegraphics[width=\linewidth]{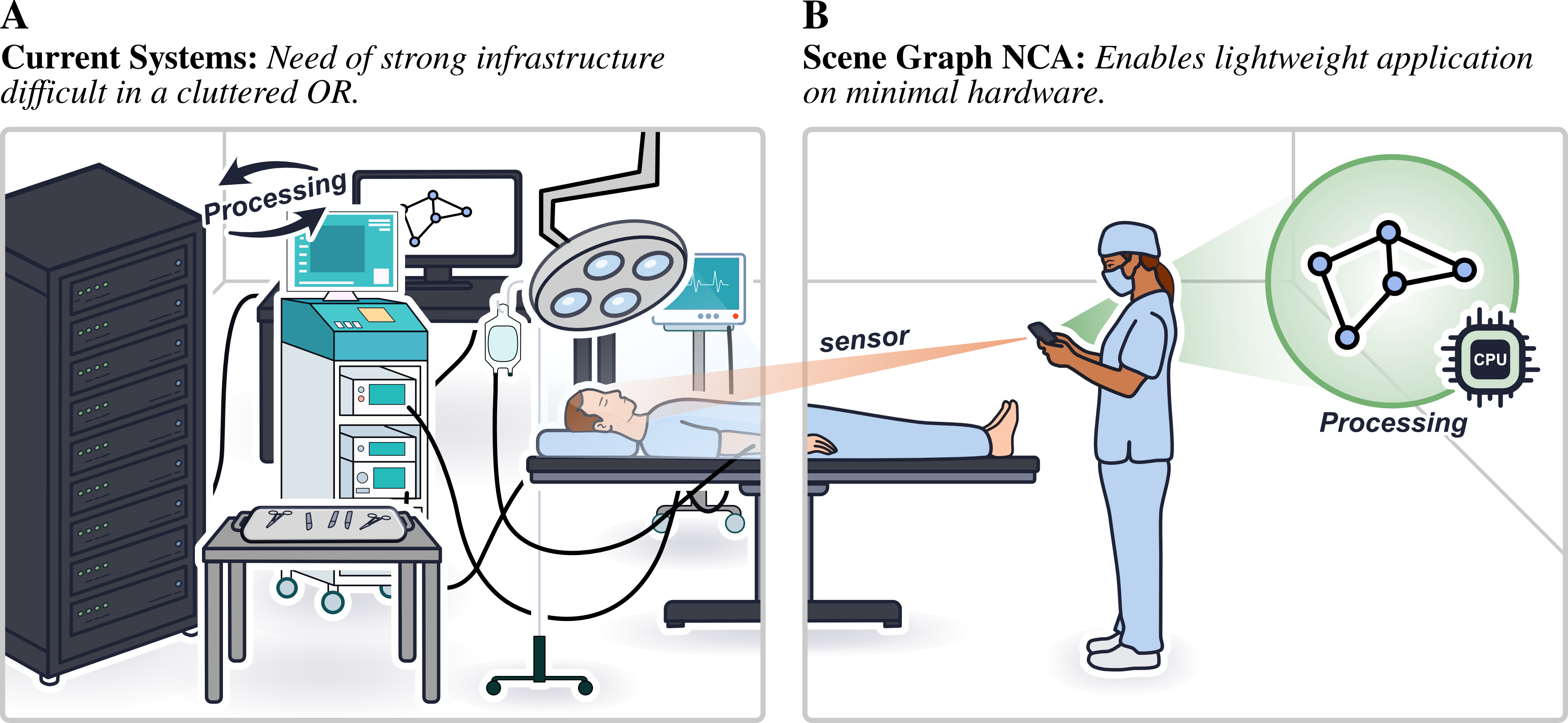}
    \caption{Lightweight scene graph generation on fanless devices such as smartphones.}
    \label{fig:intro_figure}
\end{figure}

Scene graph generation from surgical videos is leading to a holistic understanding of the surgery, especially structured representations of the scene semantics relationships and interactions~\cite{henriques2025decoding}.
State-of-the-art (SOTA) scene graph generation relies on huge deep learning models consisting of millions of parameters~\cite{ozsoy2024oracle}, which demand massive workstations for deployment in the operating room (OR). 
However, such workstations suffer from practical concerns, such as: 1) Hygiene: Workstations are difficult to sterilize, and the fans distribute dirt across the OR~\cite{world2016global}. 
2) Footprint: Big workstations further reduce the already narrow space in the OR, and tethered connections add logistical complexity.
Utilizing cloud computing is not feasible either, as this demands a stable internet connectivity and introduces high latency and data sovereignty issues. 
A lightweight scene graph generation alternative that infers on edge devices provides an affordable, responsive, and secure solution that 1) runs on fanless, sealed machines, which are easily sterilizable, 2) keeps data in the OR, offering privacy-by-design, and 3) democratizes access by leveraging ubiquitous hardware (Fig.~\ref{fig:intro_figure}).

%Other recent scene graph generation algorithms use large models, such as vision transformers or CNNs, for detection and feature extraction.
%Those algorithms are capable of producing high-quality scene graphs, but demand large computing resources. 
SOTA scene graph generation tailored for the clinical setting uses large vision language models~\cite{ozsoy2024oracle}, vision transformers~\cite{pei2024s}, or convolutional neural networks~\cite{sanner2024voxel}.
All of those methods are parameter-heavy, rendering the proposed methods impractical for clinical deployment on low-power fanless devices.
Neural Cellular Automata (NCA), on the other hand, are lightweight deep learning models,  well-suited for medical applications.
NCAs have previously been used for binary segmentation of single anatomies on modalities like MRI~\cite{kalkhof2023med,kalkhof2023m3d}, X-Ray~\cite{kalkhof2024unsupervised}, and ultrasound~\cite{lemke2025equitable}.
To the best of our knowledge, only one NCA~\cite{lemke2025octreenca} has been trained for multi-class segmentation; however, no previous NCA has been trained for learning structured representations, such as scene graphs.
%However, none of those NCAs were trained to generate a holistic understanding of the anatomies present in the image.
%To the best of our knowledge, only one NCA has been applied to surgical data~\cite{lemke2025octreenca}, which segmented only 5 classes in the cholecystectomy surgery, failing at generating a holistic, let alone symbolic, representation of the surgical procedure.

%Previously, NCAs have been used for a variety of tasks in medical imaging, including segmentation~\cite{kalkhof2023med,kalkhof2023m3d}, registration~\cite{ranem2024nca}, and generation~\cite{kalkhof2025parameter,lang2025temporal}.
%However, to the best of our knowledge, NCAs have never been used for learning representations, let alone for learning symbolic representations such as scene graphs.  

% What
% How

We design our scene graph generation algorithm, \textbf{SG-NCA}, based on NCAs combined with an octree data structure~\cite{lemke2025octreenca} for efficient object detection and feature retrieval from surgical videos.
We design a class-incremental curriculum for efficient NCA training, tailored for the complex clinical setting. 
%The NCA is trained according to a class curriculum, tailored for the complex clinical setting. 
After the NCA segments the anatomies and tools in the frames, graph nodes and geometric relations are automatically inferred from the segmentation mask.
Finally, our parameter-efficient segmentation-grounded~\cite{khandelwal2021segmentation} relation predictor infers semantic relations based on the features produced by the segmentation NCA.

Our contributions are as follows: 
1) We propose the first scene graph generation algorithm based on NCA, the first NCA for structured representation learning.
2) We evaluate our algorithm on two video recordings of cataract surgery and cholecystectomy, showing that SG-NCA holds up to established baselines while requiring $55\times$ fewer parameters.
3) We deploy our model on edge devices that run within the thermal and hygiene constraints of the OR and automatically generate captions for surgical scenes right on the edge of the bedside.

\section{Methodology}
We describe our segmentation-grounded scene graph generation algorithm. First, we briefly introduce the NCA-based scene graph generation. 
After that, we outline the NCA segmentation model and, finally, we elucidate how we adapt the NCA to the surgical setting.
\begin{figure}[t]
    \centering
    \includegraphics[width=\linewidth]{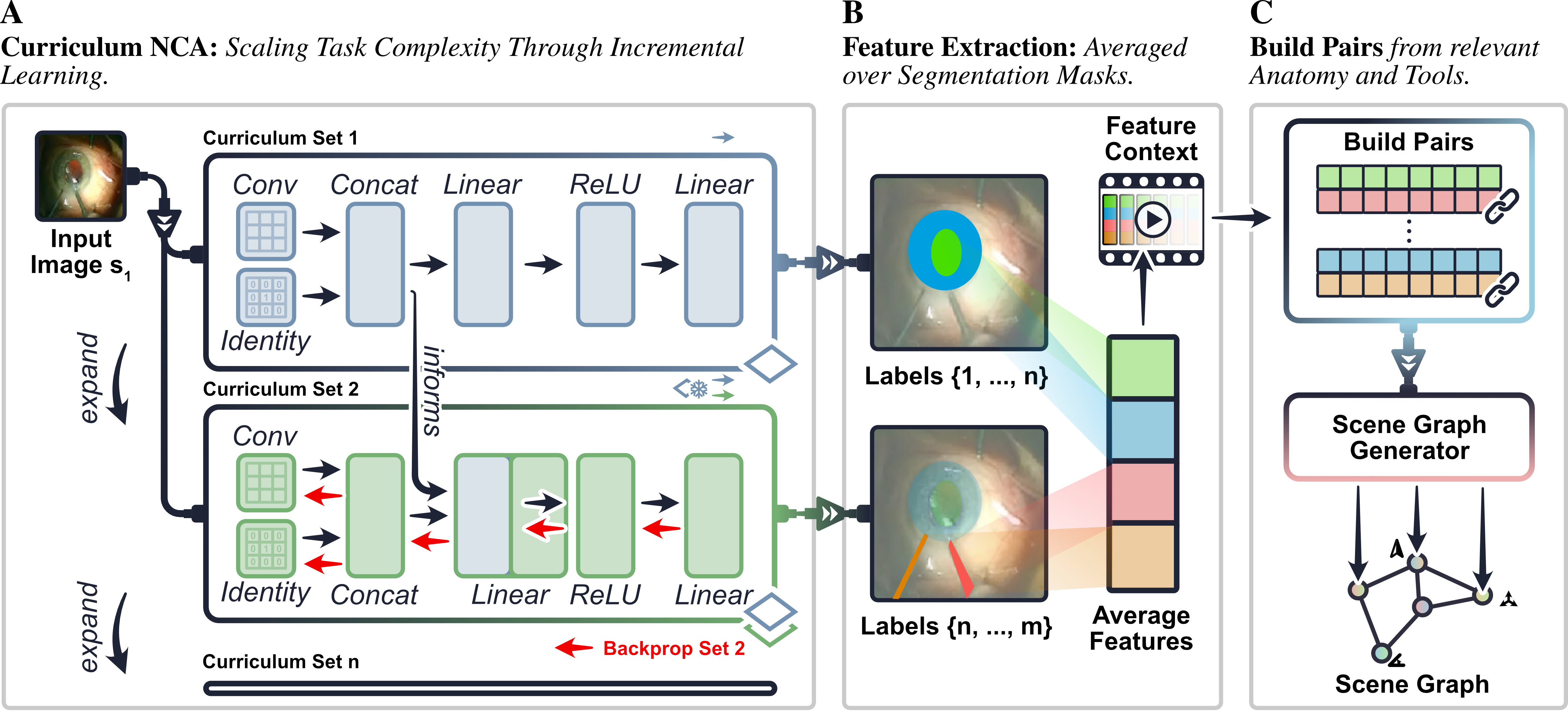}
    \caption{The curriculum-based training (A), the segmentation-grounded feature extraction (B) and the relation classification (C).}
    \label{fig:method}
\end{figure}

\subsection{Segmentation-Grounded Scene Graph Generation}
We infer scene graphs from the probability distribution $p(G|I)$ of the scene graph $G$ conditioned on the image $I$. Since learning this distribution is difficult, we decompose it into
$$
p(G|I) = p(O|I) \cdot p(R|O,I)
$$
where $O$ are the objects, and $R$ are their relationships.
The objects $O$ are inferred by segmenting tools and anatomies in the given frame with the NCA.
In the second stage, we infer semantic relationships (e.g. \textit{retracting}, \textit{holding}, \textit{inserting}) from $p(R|O,I)$, which includes the surgery-specific prior. 

Algorithm~\ref{alg:method} describes our video scene graph generation in pseudo-code.
First, SG-NCA $f_\theta$ generates the segmentation mask $S$ and the features $Z$ for the latest frame $I_t$, as outlined in Sec.~\ref{sec:segmentation}.
Segmentation masks that are larger than a pre-defined threshold $\tau=150$ pixels constitute nodes $O_t$ in the graph.
Taking inspiration from segmentation-grounded scene graph generation~\cite{khandelwal2021segmentation}, we average the pixel-wise features $z_c$ corresponding to each object $o_c$ (Fig.~\ref{fig:method} B).
Since object features are high-dimensional, we learn a projection matrix that embeds object features in a 64-dimensional embedding space.
A second projection mechanism fuses the features with those of 7 previous frames within a 1-second window stored in the cache $C$, and projects them to a temporally-enriched 256-dimensional feature vector for each vector.
Based on those features, the 3-layer relation classifier $g_\theta$ predicts the semantic relationships $r_{ij}$ for all possible pairs $(o_i,o_j)$, taking the data-specific prior $p(R)$ into consideration (Fig.~\ref{fig:method} C).
Finally, the geometric \textit{close to} relation is inferred from objects with touching segmentation masks.

\begin{algorithm}[t]
\caption{Scene Graph Generation with SG-NCA}
\label{alg:method}
\begin{algorithmic}[1]

\State \textbf{Inputs:}
\State \quad $I_t$ : image at time $t$
\State \quad $f_\theta$ : segmentation + feature NCA
\State \quad $g_\theta$ : relation classifier
\State \quad $C$ : object feature cache from previous frames

\State \textbf{Output:}
\State \quad $G_t = (O_t, R_t)$ : scene graph with objects $O_t$ and relations $R_t$
\State \quad $C$ : updated cache for subsequent frames

\vspace{2mm}

\Function{UpdateSceneGraph}{$I_t, C$}

    \State $S, Z \gets f_\theta(I_t)$
    \Comment{segmentation labels and features}

    \State $O_t \gets \emptyset$ \Comment{initialize objects}

    \For{each class $c$ in $S$}
        \State $M_c \gets \{p \mid S(p)=c\}$ \Comment{segmentation mask}
        \If{$|M_c| > \tau$}
            \State $z_c \gets \textsc{Pool}(Z, M_c)$ \Comment{segmentation-grounding}
            \State $z_c \gets \textsc{TemporalFuse}(z_c, c, C)$
            \State add node $o_c = (c, z_c)$ to $O_t$
        \EndIf
    \EndFor

    \State $R_t \gets \emptyset$ \Comment{initialize relations}

    \For{each ordered pair $(o_i, o_j) \in O_t$}
        \State $r_{ij} \gets g_\theta(z_i, z_j)$ \Comment{predict relation from feature pair}
        \State add edge $(o_i, r_{ij}, o_j)$ to $R_t$
    \EndFor

    \State $C \gets \textsc{UpdateCache}(C, O_t)$

    \State \Return $(O_t, R_t), C$

\EndFunction
\end{algorithmic}
\end{algorithm}

\subsection{NCA for High-resolution Scene Graph Generation}\label{sec:segmentation}
NCAs are lightweight segmentation and feature extraction models inspired by cellular automata such as Conway's Game of Life.
However, instead of hand-engineered update rules, the NCA uses a neural network to learn its update rule.
%Due to its local-only communication within a $3\times3$ grid, NCAs struggle to diffuse information across the entire image.
The recently proposed OctreeNCA~\cite{lemke2025octreenca} generalizes the neighborhood definition by embedding the image in an octree data structure for efficient knowledge diffusion on a coarse scale, and fine-grained segmentation on a fine scale.
The OctreeNCA downscales the input image to a $\frac{1}{2^5}$ of its original resolution and diffuses global knowledge using the first NCA.
After that, the hidden states are upscaled by $2\times$ and concatenated with the next-finer scale of the image in the next octree level.
The procedure is repeated until the final NCA delivers the segmentation masks.
The other segmentation logits and the other states from all octree levels (including the last) are concatenated and used for relation prediction.

%Due to its repetitive nature, previous works have successfully employed NCA for medical image segmentation~\cite{kalkhof2023m3d,kalkhof2023med}.

%The NCA is applied to a regular lattice of which each cell has $C=20$ dimensions.
%The RGB channels of the input image are embedded into the first three channels of the lattice. The NCA then iterates over the grid to form the segmentation in channels behind the input. 
%The remaining channels (hidden states) can be used by the NCA to store learned information for an informed segmentation. 
%Since NCAs inherit the local-neighborhood constraint of cellular automata, the traversal of information across the scale of the image is slow and inefficient, especially if the image is of high resolution.  
%To enable the efficient transfer of global information while keeping small details in the segmentation mask, we use OctreeNCA~\cite{lemke2025octreenca} with five levels.

\subsection{Curriculum NCA Training for Many Classes}
As the only way NCAs can emit segmentation masks is within their cellular grid, NCAs are inherently constrained in the number of classes they can learn to segment.
Assuming the NCA has a $C$-dimensional input lattice, and $3$ input channels (RGB), the NCA can segment at most $C-3$ classes.
%In surgical data, the number of anatomies and tools can reach 17.
%In the case of an NCA with $C=20$, there would be no hidden states for the NCA to store auxiliary information, decreasing the NCA's ability to produce high-quality segmentation masks. 
Simply increasing the number of dimensions $C$ increases computational demand and, due to the repetitive nature of NCA, scales very poorly in terms of computational requirements during training.

Instead, we propose an efficient class-curriculum learning algorithm for NCAs by introducing classes in small batches.
Our SG-NCA first establishes a basic understanding of the surgical scene by training on 5 of the most frequent classes. 
After that, the dimension $C_0$ of the hidden states is extended by $C_+$ to account for the new classes and additional hidden states $C_{\task+1}= C_\task+C_+$.
Since old parameters are frozen, only the lightweight set of new parameters must be trained.
This significantly reduces the size of the computational graph needed for backpropagation, rendering multi-class training feasible and efficient (Fig.~\ref{fig:method}~A).

%Figure~\ref{fig:computational_graph} shows the computational graph of one update step during SG-NCA training at stage $t+1$. The cells in the cellular grid are denoted as $\vec a_i\in\R^{C_t}$ and $\vec a_i\in\R^{3}$, $\Phi$ and $\Psi$ compute the depth-wise convolution, and $A,B,C,D$ are matrices for computing the additive update.
%$B$ uses the information learned in the previous stage of SG-NCA as well as the information stored in the newly added channels by stacking $[\Psi(\vec{a}), \vec a, \Phi(\vec b), \vec b]$.
%Since only the added parameters need to be learned, the backward pass is only required for a small subset of parameters, rendering training efficient.

\section{Experimental Setup}
In this section, we highlight the data used in our study and the baseline algorithms for scene graph generation.

%\subsection{Data}
\textbf{Cholecystectomy: }For our experiments on cholecystectomy, we leverage videos of the Cholec80 dataset~\cite{endonet}. It contains videos captured at 25 FPS of 80 patients.
The CholecSeg-8k dataset~\cite{hong2020cholecseg8k} is a subset of Cholec80 containing roughly 8,000 frames with dense segmentation masks.
We use CholecT50~\cite{nwoye2020recognition}, which is annotated with action triplets, for training and evaluation of the scene graph generation. 
% We align subject and object classes with the segmentation data, and  
%Since annotated objects between CholecSeg-8k, and CholecT50, we remap relations such that both subject and object are annotated classes in CholecSeg-8k. Further, we remove classes that cannot be mapped. The final scenario consists of XX classes and XX relation labels.

\textbf{Cataract surgery: }We conduct experiments on the CATARACTS dataset~\cite{al2019cataracts}, which comprises 50 videos of surgeons performing cataract surgery recorded at 30 FPS. Since this data does not contain segmentation labels, we evaluate on Cadis~\cite{grammatikopoulou2021cadis}, which is a subset containing dense segmentation labels. For training and evaluation of our scene graph generation, we use the CAT-SG dataset~\cite{holm2025cat}.
%Since some labels of CAT-SG are not present in the Cadis annotations, we remap and remove some relations similar to those we have done in the Cholecystectomy data. Finally, we receive XX classes and XX relation labels.

Since segmentation annotations are scarce for both cases, we leverage pseudo-masks generated from SASVi~\cite{sivakumar2025sasvi}, which relies on SAM2~\cite{ravi2024sam} augmented with an automated prompting network.
For evaluating the segmentation performance of our method, we ensure all ground-truth segmentation masks are in the validation and test split. The remaining cases are split randomly.
For both domains, we ensure a consistent patient split between training, validation, and test data.

%\subsection{Evaluation}
\textbf{Evaluation: }%
We evaluate the segmentation models using the Dice score, which measures the overlap of the predicted and the true segmentation mask.

The scene graph generation is evaluated using the unconstrained Recall@K, mRecall@K, and mAP@K metrics.
Since there can be up to 3 relations at once in the Cholecystectomy data, we use $K=4$ for this data and $K=6$ for the Cataracts data, as there can be up to 5 relations in a single frame.
Our metrics do not impose graph constraints, meaning one pair of objects can have multiple relationships, e.g., the grasper grasping and retracting at the same time.
% https://github.com/KaihuaTang/Scene-Graph-Benchmark.pytorch/blob/master/METRICS.md#explanation-of-our-metrics

%\subsection{Baselines}
\textbf{Baselines: }%
We implement several segmentation baselines and combine them with the MotifNet~\cite{zellers2018neural} relation prediction network. 
MotifNet uses a biLSTM to transfer knowledge between objects.
A final linear layer predicts the relations based on the enriched features. 
For segmentation, we use UNet~\cite{ronneberger2015u}, which is a fully convolutional network, and SegFormer~\cite{xie2021segformer} and SwinUNet~\cite{cao2022swin}, which are transformer-based architectures.
We replace the Swin transformer layers with more efficient variants from SwinV2~\cite{liu2022swin}.
We develop a parameter-efficient UNet variant, which we refer to as tinyUNet.

\section{Results}
\begin{table}[t]
    \caption{Segmentation results and number of parameters for each segmentation model.}
    \centering
    \begin{tabular}{l|rrr|rrr}
\toprule
 & \multicolumn{3}{c|}{Cholecystectomy} & \multicolumn{3}{c}{Cataracts} \\
 & macro Dice & micro Dice & \#Params & macro Dice & micro Dice & \#Params \\
\midrule
\textbf{SG-NCA} & \sbest70.9 $\pm$ 16.4 & 76.6 $\pm$ 21.9 & \best \textbf{27,465} & \best75.0 $\pm$ 15.9 & \sbest 82.1 $\pm$ 18.0 & \best\textbf{42,505} \\
SegFormer & \best71.3 $\pm$ 28.4 & \best81.7 $\pm$ 22.1 & \sbest 3,717,484 & \sbest 70.8 $\pm$ 21.2 & \best 82.9 $\pm$ 20.5 & \sbest3,719,283 \\
UNet & 44.4 $\pm$ 25.7 & 53.6 $\pm$ 25.2 & 68,331,670 & 43.1 $\pm$ 29.5 & 63.7 $\pm$ 25.9 & 68,332,580 \\
SwinUNetv2 & 67.4 $\pm$ 27.9 & \sbest78.0 $\pm$ 23.2 & 27,941,028 & 67.4 $\pm$ 20.2 & 79.0 $\pm$ 20.8 & 27,941,700 \\
\bottomrule
\end{tabular}
    \label{tab:seg_results}
\end{table}
In this section, we evaluate SG-NCA's segmentation and scene graph generation capabilities, and present an ablation study demonstrating the robustness of our algorithm.
Finally, we show video captions generated by SG-NCA and compare different edge devices with a clinical workstation for deployment.

\textbf{Segmentation: }%
Segmentation accuracy has a significant impact on relation prediction, as objects must be localized accurately, and low-quality segmentations can degrade downstream performance~\cite{sanner2024voxel}.
Table~\ref{tab:seg_results} reports the segmentation results, together with the number of parameters.
Due to the class-curriculum training, our SG-NCA achieves good Dice scores on all classes, leading to an overall high macro Dice score.
SG-NCA performs slightly worse than the SOTA in delineating the common anatomies, while requiring $87\times$ fewer parameters.
Overall, our segmentation backbone requires less than $1.2\%$ of the parameters of its baselines.

\textbf{Scene Graph Generation: }%
\begin{figure}[t]
    \centering
    \includegraphics[width=\linewidth]{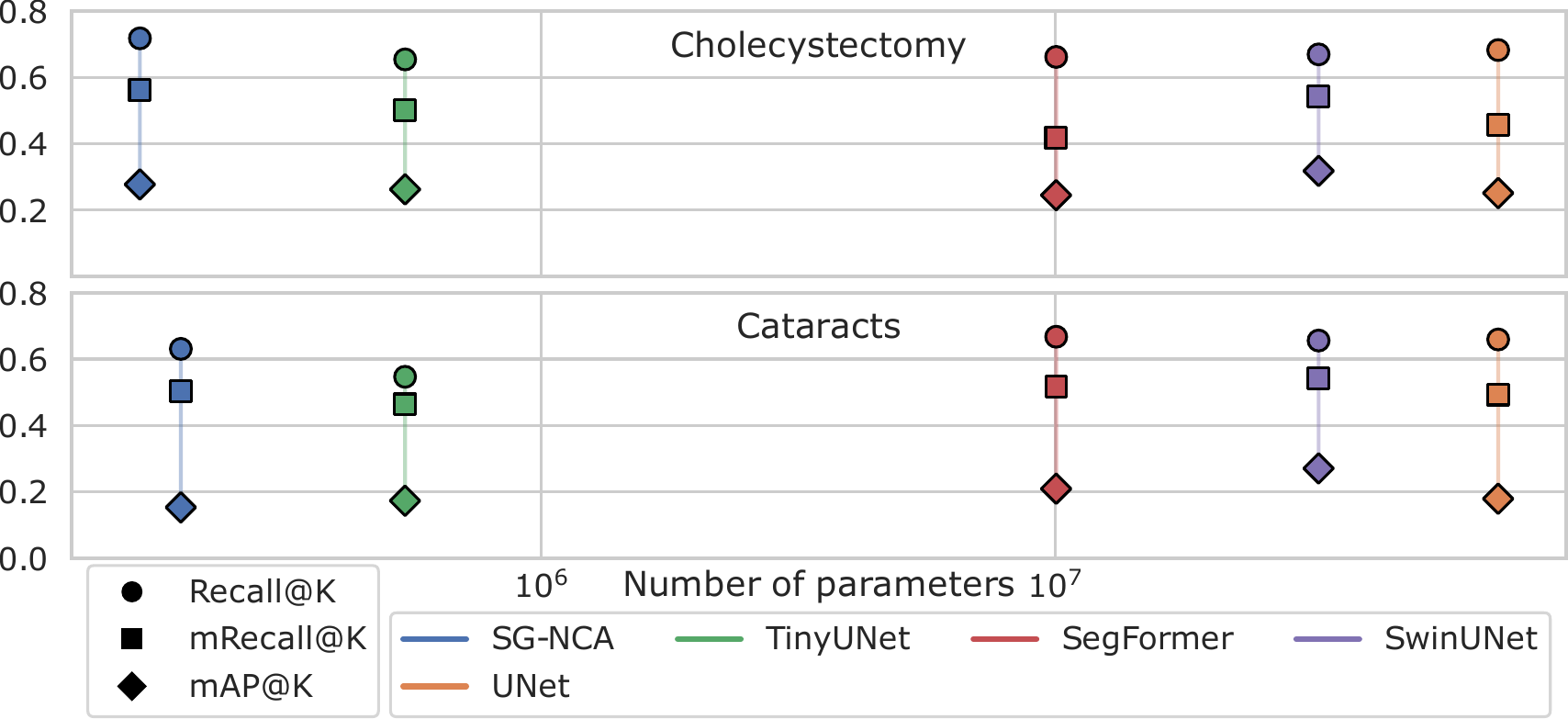}
    \caption{Scene graph generation evaluation of semantic relationships. SG-NCA requires $55\times$ fewer parameters while achieving comparable performance, running on an edge device.}
    \label{fig:sgg_results}
\end{figure}
Figure~\ref{fig:sgg_results} reports the scene graph generation results of SG-NCA and its baselines.
Our SG-NCA achieves similar results to its baselines, while requiring $55\times$ fewer parameters than the most lightweight baseline SegFormer.
Our method's lightweight design allows inference right on the edge without requiring a GPU.

% tables side-by-side: https://arxiv.org/abs/2303.13293

\textbf{Ablation Study: }%
\begin{table}[t]
    \caption{Ablation on the increments of channels $C_+$ and hidden size $H_+$ during SG-NCA training.}
    \centering
    \setlength{\tabcolsep}{3pt}
\begin{tabular}{ll|rrr|rrr}
\toprule
 & & \multicolumn{3}{c|}{Cholecystectomy} & \multicolumn{3}{c}{Cataracts} \\
$C_+$ & $H_+$ & macro Dice & micro Dice & \#Params & macro Dice & micro Dice & \#Params \\
\midrule
\multirow[c]{3}{*}{8} & 32 & 73.7 $\pm$ 11.7 & 77.4 $\pm$ 20.8 & 59,625 & 75.9 $\pm$ 13.6 & 81.7 $\pm$ 17.7 & 148,905 \\
 & 16 & 73.9 $\pm$ 12.0 & 77.6 $\pm$ 21.4 & 40,185 & 76.4 $\pm$ 12.7 & \best 82.6 $\pm$ 17.6 & 85,625 \\
 & 8 & \best74.3 $\pm$ 11.9 & \best77.7 $\pm$ 20.7 & 30,465 & 75.2 $\pm$ 13.6 & 81.6 $\pm$ 18.4 & 53,985 \\
\midrule
\multirow[c]{2}{*}{4} & 16 & 74.0 $\pm$ 12.0 & 77.6 $\pm$ 21.0 & 34,785 & \best 78.0 $\pm$ 09.7  & 82.5 $\pm$ 17.6 & 55,785 \\
 & 8 & 72.6 $\pm$ 13.1 & 76.8 $\pm$ 21.6 & \best 27,465 & 75.6 $\pm$ 14.1 & 82.1 $\pm$ 17.8 & \best42,505 \\
\bottomrule
\end{tabular}
    \label{tab:dice_ablation}
\end{table}
We report results of our ablation study on the curriculum-based segmentation training in Tab.~\ref{tab:dice_ablation}.
Essentially, the increment of the number of channels $C_+$ and the corresponding hidden size $H_+$ has minimal influence on the segmentation performance.
Even our smallest configuration with very small increments of $C_+=4$ channels and $H_+=8$ maintains reasonable segmentation performance.
Hence, we select this configuration for our experiments on scene graph generation.  

\textbf{Caption Generation: }%
\begin{figure}[t]
    \centering
    \includegraphics[width=\linewidth]{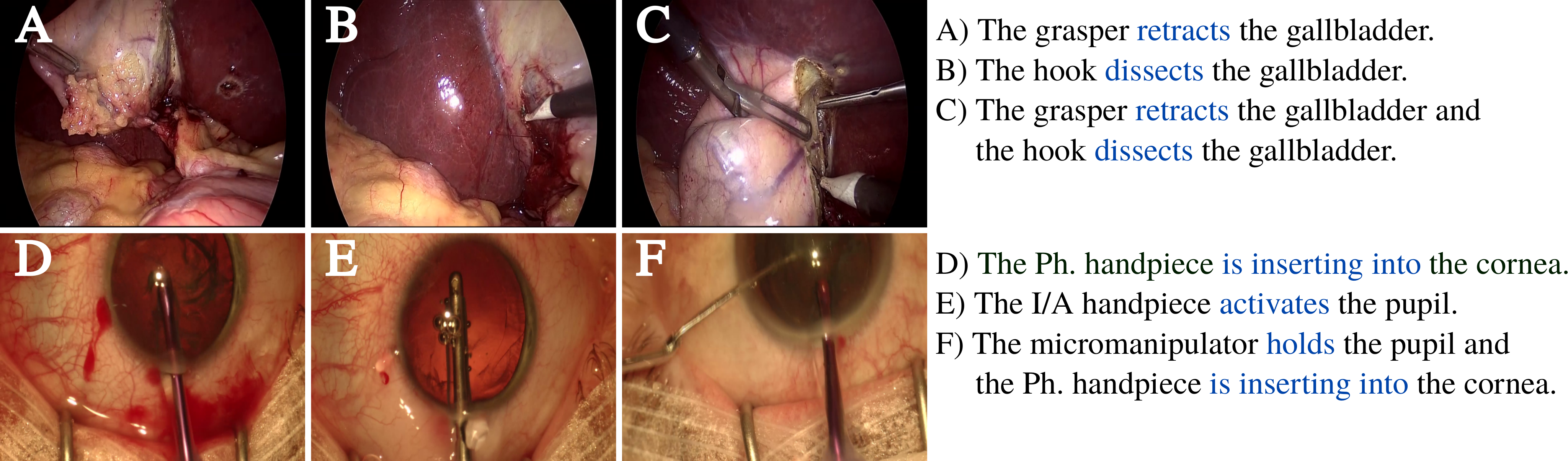}
    \caption{Captions generated from SG-NCA. See supplementary material for videos.}
    \label{fig:qualitative_results}
\end{figure}
Based on the inferred scene graphs, we demonstrate rule-based caption generation, describing the workflow of the surgery. 
Figure~\ref{fig:qualitative_results} shows examples of those captions.
Videos augmented with scene graphs and captions can be found in the supplementary material. The \textit{close to} relations are indicated by thin lines, whereas bold green ones indicate semantic relations.

\begin{table}[t]
    \caption{Maximum Memory demand (in MB), Temperature increase (in \textdegree K), and Energy consumption (in W) during inference of SG-NCA on various hardware. \textcolor{red}{Due to thermal and power requirements, other models exceed the OR-limits.}}
    \centering
    \setlength{\tabcolsep}{3pt}
\begin{tabular}{l|c|cc|cc|cc}
\toprule
& &\multicolumn{2}{c|}{\hot Workstation} & \multicolumn{2}{c|}{Smartphone} & \multicolumn{2}{c}{Raspberry Pi} \\
& Mem. & \hot Temp. & \hot Energy & Temp. & Energy & Temp. & Energy\\
\midrule
SG-NCA & 44.25& \hot 4.18 & \hot 223 & 0.66& 1.6 & 0.62 & 4.5\\
\bottomrule
\end{tabular}
    \label{tab:temperature}
\end{table}
\textbf{Deployment on the Edge: }%
Next to the large workstation, we deploy and benchmark our model on a smartphone and a Raspberry Pi, both of which are low-energy computing devices. 
In Tab.~\ref{tab:temperature}, we report the room temperature increase after 40 minutes of runtime, and the average power draw during inference on all three devices.
The smartphone and the Raspberry Pi are both developed for minimal energy consumption and hence have very little impact on the temperature of the room, whereas the workstation PC significantly heats the room, while contaminating the room with its fans.

\section{Conclusion}
We propose SG-NCA, a lightweight model for scene graph generation using NCAs.
The proposed curriculum-based training enables training large NCAs with minimal computational overhead, effectively equipping them with the ability to segment a wide range of surgical anatomies and tools.
Combined with the lightweight relation classifier, SG-NCA generates scene graphs without increasing computational demand.

Our experiments show that SG-NCA competes with models that are much larger in terms of scene graph generation and segmentation performance.
Hence, our model can run on fanless, easily sanitizable hardware, as required by operating room hygiene standards.
SG-NCA enables a holistic understanding of surgery at the edge of the bedside.

\begin{credits}
\subsubsection{\ackname}
This work has been partially funded by the Federal Ministry of Research, Technology and Space project ``Advice'' (grant 13GW0817C).

\subsubsection{\discintname}
The authors have no competing interests to declare that are relevant to the content of this article.
\end{credits}

\bibliographystyle{unsrt}
\bibliography{references}
\end{document}